\documentclass[lettersize,journal]{IEEEtran}
\usepackage{amsmath,amsfonts,amssymb,amsthm} 
\usepackage{algorithmic}
\usepackage{algorithm}
\usepackage{array}
\usepackage[caption=false,font=normalsize,labelfont=sf,textfont=sf]{subfig}  
\usepackage{textcomp}
\usepackage{stfloats}
\usepackage{url}
\usepackage{verbatim}
\usepackage{cite}

\usepackage{graphicx} 
\usepackage{booktabs}
\usepackage{multirow}
\usepackage{makecell}
\usepackage{diagbox}  

\usepackage[utf8]{inputenc}

\graphicspath{{figures/}}

\newtheorem{lemma}{Lemma}
\newtheorem{theorem}{Theorem}

\begin{document}

\title{HWL-HIN: A Hypergraph-Level Hypergraph Isomorphism \\ Network  as Powerful as the Hypergraph Weisfeiler-Lehman Test with Application to Higher-Order Network Robustness}

\author{Chengyu Tian, Wenbin Pei}

\markboth{}%
{Tian \MakeLowercase{\textit{et al.}}: HWL-HIN: A Hypergraph-Level Hypergraph Isomorphism \\ Network  as Powerful as the Hypergraph Weisfeiler-Lehman Test with Application to Higher-Order Network Robustness}

\maketitle

\begin{abstract}
Robustness in complex systems is of significant engineering and economic importance. However, conventional attack-based \textit{a posteriori} robustness assessments incur prohibitive computational overhead. Recently, deep learning methods, such as Convolutional Neural Networks (CNNs) and Graph Neural Networks (GNNs), have been widely employed as surrogates for rapid robustness prediction. Nevertheless, these methods neglect the complex higher-order correlations prevalent in real-world systems, which are naturally modeled as hypergraphs. Although Hypergraph Neural Networks (HGNNs) have been widely adopted for hypergraph learning, their topological expressive power has not yet reached the theoretical upper bound. To address this limitation, inspired by Graph Isomorphism Networks, this paper proposes a hypergraph-level Hypergraph Isomorphism Network framework. Theoretically, this approach is proven to possess an expressive power strictly equivalent to the Hypergraph Weisfeiler-Lehman test and is applied to predict hypergraph robustness. Experimental results demonstrate that while maintaining superior efficiency in training and prediction, the proposed method not only outperforms existing graph-based models but also significantly surpasses conventional HGNNs in tasks that prioritize topological structure representation.
\end{abstract}


\section{Introduction}
\IEEEPARstart{C}{omplex}  systems in the real world, ranging from academic collaborations \cite{patania2017shape} and biological metabolic pathways \cite{battiston2020networks} to social communication networks \cite{benson2016higher}, are fundamentally characterized by group-wise interactions involving multiple entities \cite{patania2017shape,lambiotte2021modularity}. The robustness of complex systems refers to their fundamental ability to sustain core functionality despite external disturbances or internal failures \cite{roberts1980robustness,meena2023emergent}. It plays a crucial role in numerous fields \cite{grassia2021machine,lou2023structural} such as power systems \cite{dobson2007complex}, transportation networks \cite{liu2023port} and the internet \cite{sarkar2016internet}. Therefore, assessing robustness is crucial for understanding and enhancing complex systems. However, widely applied posterior metrics \cite{lou2023structural} based on attack simulations often incur significant computational overhead \cite{chan2016optimizing,freitas2022graph}. To address this issue, numerous studies on robustness prediction have emerged to accelerate computations.

Traditional machine learning algorithms, such as K-Nearest Neighbors \cite{mucherino2009k}, linear regression \cite{james2013introduction}, decision trees, and random forest algorithms \cite{quinlan1986induction}, have been demonstrated to effectively predict robustness \cite{dhiman2020using}. However, these methods typically necessitate the preprocessing of graph topological features, exhibit limited expressiveness for complex structures, often fail to meet stringent accuracy requirements, and lack scalability for large-scale graph data. Subsequently, graph embedding techniques were introduced to compress graphs into low-dimensional vectors \cite{wang2019surrogate}. While enabling automated feature extraction, this non-end-to-end paradigm frequently results in the loss of valuable topological information during the compression process. To capture local patterns, other approaches treated networks as images, applying convolutional operations via CNNs with iterative optimization \cite{lou2021convolutional,lou2022learning,wu2023spp}. Although this strategy enhances expressiveness by preserving features from the input adjacency matrix, it falls short in addressing the non-Euclidean nature of graphs; CNNs inherently lack the capability to capture complex topological structures and suffer from computational inefficiency when processing large-scale graph data. Building upon this, graph-native deep learning approaches emerged. Spatial graph convolutional networks \cite{zhang2024network} and self-attention mechanisms \cite{zhang2024graph} were employed for prediction, while powerful graph isomorphism networks were utilized to simultaneously predict multiple metrics via multi-task learning \cite{wu2024multitask}. These approaches have significantly enhanced prediction performance.

Despite these advancements, two critical limitations remain in the aforementioned studies. First, they largely overlook the dynamic nature of cascading failures in complex networks, where the failure of a single node can trigger a chain reaction of subsequent collapses. Second, and more fundamentally, these methods are strictly predicated on pairwise graph structures. Although pairwise graphs have served as the de facto standard for modeling complex systems \cite{newman2010networks,albert2002statistical}, abstracting interactions as pairwise links between nodes, this pairwise paradigm is inherently insufficient for representing the higher-order relationships prevalent in real-world networks \cite{zhang2024graph}. This renders traditional graph-based methods inadequate for capturing the full topological complexity required for accurate robustness assessment. To overcome this limitation, hypergraphs have been widely introduced as a generalized topological framework for precisely modeling these complex higher-order relationships \cite{zhou2006learning,antelmi2023survey}. To address the challenges posed by higher-order topological relationships, early approaches on hypergraph learning primarily adopted expansion strategies to bridge the gap between hypergraphs and standard graph algorithms. Typical methods, such as clique expansion and star expansion \cite{zhou2006learning, agarwal2006higher}, convert a hypergraph into a weighted clique graph or a bipartite graph, respectively. These transformations allow the direct application of mature GNNs or CNNs to hypergraph data. However, this pairwise paradigm inherently suffers from structural distortion and incurs prohibitive spatial and computational overhead \cite{yadati2019hypergcn, huang2021unignn}.

To mitigate the information loss caused by expansions, native HGNNs have been introduced to operate directly on the hypergraph structure. Pioneering works, such as HGNNs \cite{feng2019hypergraph}, HyperGCN \cite{yadati2019hypergcn}, and HNHN \cite{dong2020hnhn}, proposed spectral convolution and spatial message passing mechanisms tailored for hypergraphs. These methods perform end-to-end learning without intermediate conversion, capturing higher-order correlations more effectively than expansion-based methods. Nevertheless, most existing HGNNs employ simple aggregation functions (e.g., mean or max pooling) to combine features from incident vertices or hyperedges. However, the topological expressive power of these aggregators has not yet reached the theoretical upper bound required for hypergraph-level regression tasks. Similar to how standard GCNs fail to distinguish certain non-isomorphic graphs, standard HGNNs struggle to differentiate complex hypergraph topologies that exhibit subtle structural differences but similar statistical properties \cite{kim2020powerful}.

Most recently, the Weisfeiler-Lehman (WL) test has been extended to the hypergraph domain from graph \cite{feng2024hypergraph}, establishing a theoretical benchmark for the topological expressive power of hypergraph neural networks. SHGIN \cite{chen2025investigating} pioneered the integration of the hypergraph isomorphism test into neural networks, demonstrating significant efficacy in capturing higher-order structures. However, it is primarily restricted to node-level classification tasks. Furthermore, this method neglects explicit hyperedge features and fails to ensure strict injectivity during the hyperedge aggregation process, thereby theoretically limiting its capability to represent the topological structures of hypergraphs.

To address this limitation and achieve precise robustness assessment, we introduce a hypergraph-level Hypergraph Isomorphism Network framework explicitly designed for predicting higher-order network robustness. Our approach incorporates both node and hyperedge features and enforces strict injectivity throughout the dual-stage aggregation process. Subsequently, a hypergraph-level readout mechanism is employed to aggregate these enriched representations for executing hypergraph-level tasks. Specifically, the main contributions of this work are summarized as follows:

1). We propose the first hypergraph-level \textbf{H}ypergraph \textbf{I}somorphism \textbf{N}etwork (\textbf{HWL-HIN}) with an expressive power that is theoretically proven to be strictly equivalent to the \textbf{H}ypergraph \textbf{W}eisfeiler-\textbf{L}ehman test.

2). We are the first to investigate the prediction of connectivity robustness in higher-order networks, while considering scenarios involving dynamic cascading failures.

3). We conduct experiments on multiple hypergraph datasets. The results validate the superior topological expressiveness of HWL-HIN compared to standard hypergraph neural networks. Furthermore, comparative analysis demonstrates that our hypergraph-based approach consistently outperforms existing graph-based robustness prediction methods, highlighting the necessity of modeling higher-order interactions.

\section{Preliminaries}
\label{sec:preliminaries}

\subsection{Hypergraph}
A hypergraph is defined as $\mathcal{H}=(V,E)$, where $V = \{v_1,v_2,\dots, v_{|V|}\}$ is the set of vertices, and $E = \{e_1, e_2,\dots, e_{|E|}\}$ is the set of hyperedges. Each hyperedge $e\in E$ is a non-empty subset of $V$, capable of connecting multiple vertices simultaneously.

The topological structure is uniquely determined by a hypergraph incidence matrix $H \in \{0, 1\}^{|V| \times |E|}$, where $H(v, e) = 1$ if vertex $v$ is contained in hyperedge $e$ (denoted as $v \in e$), and 0 otherwise. Based on these incidence relationships, we formally define the neighbor relations and their corresponding structural properties as follows:
\begin{itemize}
    \item Vertex's hyperedge neighbors ($\mathcal{N}_e(v)$) and hyperdegree ($d_v$): The set of hyperedges that contain vertex $v$ is defined as $\mathcal{N}_e(v) = \{e \in E \mid H(v, e) = 1\}$. The hyperdegree of vertex $v$ corresponds to the size of this set, calculated as $d_v = |\mathcal{N}_e(v)| = \sum_{e \in E} H(v, e)$.
    \item Hyperedge's vertex neighbors ($\mathcal{N}_v(e)$) and cardinality ($c_e$): The set of vertices included in hyperedge $e$ is defined as $\mathcal{N}_v(e) = \{v \in V \mid H(v, e) = 1\}$. The cardinality of hyperedge $e$ corresponds to the number of vertices in this set, calculated as $c_e = |\mathcal{N}_v(e)| = \sum_{v \in V} H(v, e)$.
\end{itemize}

\subsection{Robustness Measure}
\label{subsec:robustness_def}
Network robustness quantifies the system's ability to maintain functional integrity under structural damage. We focus on \textit{connectivity robustness} against node removal attacks, which is fundamental for ensuring information flow in complex systems \cite{albert2000error}.

We adopt the percolation theory framework to evaluate robustness. The structural integrity is measured by the size of the Largest Connected Component (LCC). Conventionally, the robustness metric $R$ is defined as the discrete average of the LCC size throughout the attack sequence \cite{schneider2011mitigation}:
\begin{equation}
R = \frac{1}{N} \sum_{q=1}^{N} s(q)
\label{eq:robustness_discrete}
\end{equation}
where $s(q)$ is the fraction of nodes in the LCC after removing $q$ nodes.

However, this fixed-step discrete summation often fails to efficiently capture the critical phase transitions inherent in the percolation process, where the network structure undergoes abrupt collapse. To rigorously characterize these continuous topological features and avoid discretization bias, the robustness metric is reformulated as the definite integral of the percolation curve $s(\rho)$ over the attack fraction $\rho \in [0, 1]$ \cite{schneider2011mitigation, qu2025meta}:
\begin{equation}
R = \int_{0}^{1} s(\rho) \, d\rho
\label{eq:robustness_integral}
\end{equation}
This integral formulation provides a holistic measure of the network's functionality, implicitly emphasizing the continuous nature of the degradation process.

\subsection{Hypergraph-based Load Distribution Model}
To simulate dynamic failure propagation, the classic load distribution model is extended to higher-order systems. Similar to the load distribution models in pairwise graphs \cite{motter2002cascade}, in a hypergraph $H = (V, E)$, the initial load assigned to each node $i$ is an exponential multiple of its hyperdegree:
\begin{equation}
L_i=d_i^\beta
\end{equation}
where $\beta$ is referred to as the load index and $L_i$ represents the current load of this node. The maximum load $R_i$ that node $i$ can bear is defined as \cite{motter2002cascade}:
\begin{equation}
R_i=(1+\alpha)L_i
\end{equation}
where $\alpha>0$ is the redundancy capacity ratio, representing the tolerance margin beyond the initial load. Upon the failure of node $i$, its load is evenly redistributed among all hyperedges $e_\gamma$ to which it is connected. The amount of load allocated through each hyperedge $e_\gamma$ is:
\begin{equation}
\Delta L_{e_\gamma}=\frac{L_i}{d_i^t}
\end{equation}
where $d_i^t$ denotes the number of non-failed hyperedges associated with node $i$ at time $t$. Then, hyperedge $e_\gamma$ redistributes the load $\Delta L_{e_\gamma}$ transferred from the failed node $i$ evenly among all non-failed nodes $j$ within hyperedge $e_\gamma$:
\begin{equation}
\Delta L_j=\frac{\Delta L_{e_\gamma}}{m_\gamma^t}
\end{equation}
where $m_\gamma^t$ denotes the number of active nodes in $e_\gamma$ at time $t$.
If $L_j+\Delta L_j>R_j$, the node $j$ fails. Moreover, a hyperedge is considered failed if $m_\gamma-1$ nodes within it have failed.

\subsection{Data Generation via Adaptive Integration with Relative Precision}
Given the prohibitive cost of full-scale simulations, we argue that obtaining absolute truth for robustness labels is computationally wasteful. Since the surrogate model inevitably contains an expected prediction error $\delta_{\text{pred}}$, the true labels only require a precision margin strictly tighter than $\delta_{\text{pred}}$ to prevent label noise from dominating the training loss.

Therefore, we use a relative precision strategy derived from adaptive computational techniques \cite{qu2025meta}. We dynamically set the integration tolerance $\epsilon$ to be 50 times stricter than the target model error (i.e., $\epsilon = \delta_{\text{pred}} / 50$). This ensures that the label noise contributes negligibly to the total error budget while minimizing computational overhead. To implement this, we employ an Adaptive Simpson's method \cite{qu2025meta}.

For any given interval $[x, y]$, the Simpson's approximation $\mathcal{S}(x, y)$ is defined as the weighted sum of the endpoints and the midpoint:
\begin{equation}
\mathcal{S}(x, y) = \frac{y - x}{6} \left[ s(x) + 4s\left(\frac{x + y}{2}\right) + s(y) \right]
\label{eq:simpson_def}
\end{equation}
The algorithm recursively refines the integration by comparing a coarse estimate over $[a, b]$ against a fine-grained estimate composed of two sub-intervals $[a, m]$ and $[m, b]$ (where $m$ is the midpoint). The local truncation error $E$ is estimated by the discrepancy between these two resolutions:
\begin{equation}
E = \left| \mathcal{S}(a, b) - \left( \mathcal{S}(a, m) + \mathcal{S}(m, b) \right) \right|
\label{eq:simpson_error}
\end{equation}
If $E < \epsilon$, the fine-grained estimate is accepted; otherwise, the interval is recursively bisected to capture rapid topological transitions. To prevent infinite loops in singular regions, the maximum recursion depth is capped at $d_{\max} = 10$.

\subsection{Hypergraph Neural Networks and Hypergraph Isomorphism}
\label{sec:hgnn_iso}

\subsubsection{Hypergraph Neural Networks}
Unlike pairwise graphs, a hyperedge $e \in E$ can connect an arbitrary number of vertices, encoding higher-order correlations. 

Existing HGNNs \cite{feng2019hypergraph, yadati2019hypergcn} typically adopt a two-stage message passing paradigm. Let $h_v^{(l)}$ and $h_e^{(l)}$ denote the representations of node $v$ and hyperedge $e$ at the $l$-th layer, respectively. The general update rule is formulated as follows:

\begin{enumerate}
    \item Node-to-Hyperedge Aggregation: Each hyperedge $e$ aggregates features from its constituent nodes $V(e) = \{v \in V \mid v \in e\}$ to generate the hyperedge embedding:
    \begin{equation}
        h_e^{(l)} = \text{AGG}_{V \to E} \left( \left\{ h_v^{(l-1)} \mid v \in \mathcal{N}_v(e) \right\} \right)
        \label{eq:agg_v2e}
    \end{equation}
    
    \item Hyperedge-to-Node Aggregation: Each node $v$ aggregates features from its incident hyperedges $E(v) = \{e \in E \mid v \in e\}$ and fuses them to update its node embedding:
    \begin{equation}
        h_v^{(l)} = \text{COMBINE} \left( h_v^{(l-1)}, \text{AGG}_{E \to V} \left( \left\{ h_e^{(l)} \mid e \in \mathcal{N}_e(v) \right\} \right) \right)
        \label{eq:agg_e2v}
    \end{equation}
\end{enumerate}
Standard HGNNs predominantly utilize mean or max pooling as the $\text{AGGREGATION}$ function. 
\subsubsection{Hypergraph Isomorphism and Injectivity}
The Weisfeiler-Lehman graph isomorphism test has been extended to the hypergraph domain, termed the Hypergraph Weisfeiler-Lehman test \cite{feng2024hypergraph}. This framework establishes the theoretical upper bound for the discriminative power of hypergraph representations.

The core challenge in approximating the HWL test with neural networks lies in designing a strictly injective aggregation function. Such a function must guarantee that distinct multisets of neighbor features are mapped to unique embeddings. Drawing upon the foundational theory of Deep Sets \cite{zaheer2017deep} and Graph Isomorphism Networks (GIN) \cite{xu2018powerful}, we revisit the critical lemma governing the design of such functions:
\begin{figure}[htbp]
    \centering
    \includegraphics[width=\linewidth]{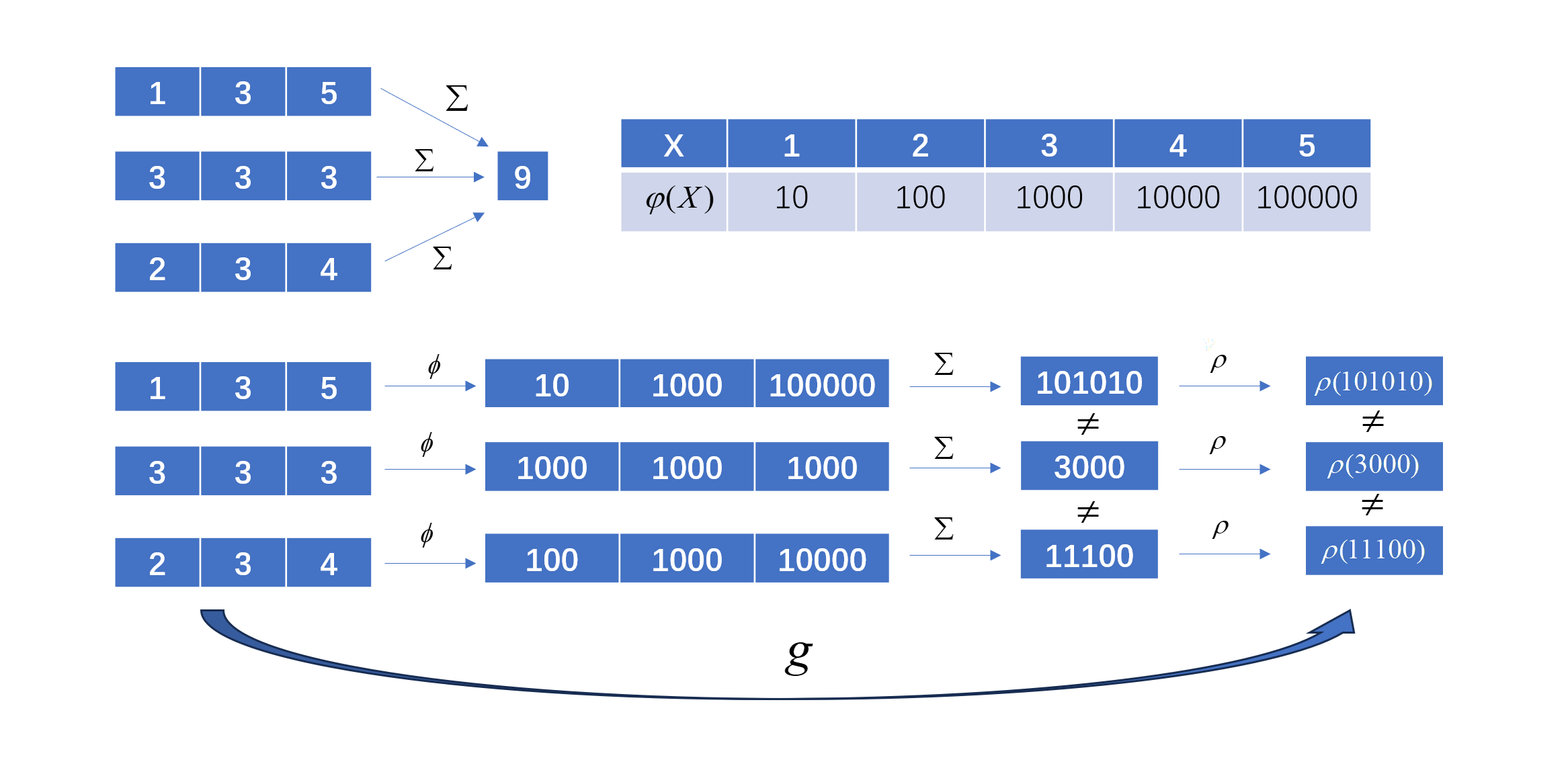} 
    \caption{Illustration of Lemma \ref{lemma:injectivity}. To ensure injectivity over multisets, element-wise features must be mapped via a non-linear mapping $\phi$ prior to summation.}
    \label{fig:map_then_sum}
\end{figure}

\begin{lemma}
\label{lemma:injectivity}
 Assume $\mathcal{X}$ is a countable set. A function $g$ defined on bounded multisets $X \subset \mathcal{X}$ is injective if and only if it can be decomposed into the form \cite{xu2018powerful}:
\begin{equation}
    g(X) = \rho \left( \sum_{x \in X} \phi(x) \right)
    \label{eq:deepsets_condition}
\end{equation}
where elements $x$ are mapped via the non-linear function $\phi$, and $\rho$ is a post-processing function.
\end{lemma}

As illustrated in Fig. \ref{fig:map_then_sum}, the application of a non-linear transformation $\phi$ to the elements is crucial for ensuring injectivity following multiset summation. In the absence of this transformation, the aggregation operation reduces to a simple linear superposition, thereby failing to satisfy the injectivity condition and consequently failing to capture the structural information required to distinguish complex higher-order topologies.

\section{Related Work}
\label{sec:related_work}

In this section, we review three distinct deep learning paradigms for network robustness prediction: Convolutional Neural Networks, Graph Transformers, and Graph Isomorphism Networks. These will be applied to bipartite graphs in the experimental section for comparative analysis.

\subsection{Convolutional Neural Networks}
\label{sec:rw_cnn}
Early CNN-based approaches treated the graph adjacency matrix as a grayscale image. However, standard CNNs require fixed-size inputs, necessitating destructive resizing for graphs of varying sizes $N$ \cite{lou2021convolutional,lou2022learning}. To address this, the SPP-CNN framework \cite{wu2023spp} introduces a Spatial Pyramid Pooling (SPP) layer to handle arbitrary graph resolutions.

First, the adjacency matrix is processed by convolutional layers to generate a feature map $F \in \mathbb{R}^{N' \times N' \times C}$. Unlike fixed-size pooling, the SPP layer extracts features at multiple scales $s \in \{1, \dots, S\}$ (e.g., $4\times4, 2\times2, 1\times1$ grids). For a given scale $s$, the pooling operation for the $(i, j)$-th bin is defined as:
\begin{equation}
    y_{s, i, j} = \max_{(p,q) \in \text{bin}(i,j)} F_{p,q,:} 
    \label{eq:spp_pooling}
\end{equation}
The outputs from all bins are flattened and concatenated to form a fixed-length vector $H_{SPP}$, which is subsequently fed into fully connected layers to predict the robustness sequence. This multi-scale aggregation allows SPP-CNN to preserve topological information without resizing artifacts.

\subsection{Graph Transformers}
\label{sec:rw_gt}
Graph Transformers capture global dependencies through self-attention mechanisms. NRL-GT \cite{zhang2024graph} employs a specialized architecture designed to integrate structural information into the attention calculation.

Specifically, it incorporates edge features $e_{ij}$ into the query-key interaction to preserve structural context. The attention coefficient $A_{ij}^{m}$ for the $m$-th head is computed as:
\begin{equation}
    A_{ij}^{m} = \text{softmax}_j \left( \frac{(W_Q^m h_i) \cdot (W_K^m h_j + W_E^m e_{ij})}{\sqrt{d_k}} \right)
    \label{eq:gt_attn_score}
\end{equation}
The model aggregates neighbor information based on these coefficients and fuses multi-head outputs with the original node features using residual connections and gating mechanisms. Finally, the node representations are updated via a Feed-Forward Network (FFN) and Layer Normalization (LN):
\begin{equation}
    h_i^{(l+1)} = \text{LN} \left( \hat{h}_i + \text{FFN}(\hat{h}_i) \right)
    \label{eq:gt_ffn}
\end{equation}
This hierarchical attention ensures the model captures both local structural details via edge-aware mechanisms and global patterns through the transformer architecture.

\subsection{Graph Isomorphism Networks}
\label{sec:rw_gin}
The GIN-MAS framework \cite{wu2024multitask} is built upon the GIN, utilizing a multitask learning approach. GIN is selected for its superior discriminative power, which is theoretically equivalent to the 1-WL test.

The node update rule employs an injective aggregation function. For layer $l$, the feature of node $v$ is updated as:
\begin{equation}
    h_v^{(l+1)} = \text{MLP}^{(l)} \left( (1 + \epsilon^{(l)}) \cdot h_v^{(l)} + \sum_{u \in \mathcal{N}(v)} h_u^{(l)} \right)
    \label{eq:gin_update}
\end{equation}
where $\epsilon^{(l)}$ is a learnable scalar. To obtain a global graph representation $H_G$, GIN-MAS employs a jumping knowledge readout that concatenates sum-pooled features from all layers:
\begin{equation}
    H_G = \text{CONCAT} \left( \sum_{v \in V} h_v^{(l)} \mid l = 0, 1, \dots, L \right)
    \label{eq:gin_readout}
\end{equation}
In Eq. \eqref{eq:gin_readout}, $L$ represents the total number of iterations. The $\text{SUM}$ aggregation over node features ensures the injectivity of the multiset function, while the $\text{CONCAT}$ operation across all $L$ layers enables the model to capture structural information at different scales, effectively forming a comprehensive graph-level embedding.

\section{Methodology}
\label{sec:methodology}
In this section, we present the HWL-HIN, an architecture provably as expressive as the Hypergraph Weisfeiler-Lehman test, designed to enable efficient and precise hypergraph robustness prediction. We first delineate the input feature representation for both nodes and hyperedges. Subsequently, we establish the theoretical conditions requisite for a hypergraph neural network to achieve maximal expressiveness. Building upon these foundations, we construct the core architecture of HWL-HIN. Finally, we introduce an adaptive learning rate scheduling strategy to enhance optimization stability and convergence performance.
\subsection{Input Feature}
\label{subsec:input}

To accelerate convergence and embed domain knowledge, we construct input features based on the structural properties of the hypergraph. Specifically, for each hyperedge $e \in E$, we define a one-dimensional feature based on its cardinality $c_e$. For each vertex $v \in V$, we construct a three-dimensional input feature vector, forming the node feature matrix $X \in \mathbb{R}^{|V| \times 3}$, where the feature vector $x_i$ for node $i$ is defined as:

\begin{equation}
x_i = [\tilde{k}_i, \tilde{c}_i, \tilde{o}_i]^\top
\label{eq:node_feat}
\end{equation}
Given that different metrics may have vastly different scales, directly using raw values can lead to training instability. To address this and ensure numerical stability during the optimization process, all input feature components are normalized to the range $[0, 1]$ via scaling. The first two features, Normalized Hyperdegree ($\tilde{k}_i$) and Normalized Local Cardinality ($\tilde{c}_i$), capture basic local connectivity. The third feature, $\tilde{o}_i$, is designed to capture the "sequential vulnerability" of nodes under a specific attack method,
serving as a structural prior for network collapse. Its definition depends on the targeted attack method:
    
    \begin{enumerate}
        \item \textit{Static Targeted Attacks:}  In static conditions, node failures caused by each attack based on hyperdegree do not affect other nodes. In such scenarios, the failure sequence is easily determined, such as by the order of node hyperdegrees from highest to lowest.
        
        \item \textit{Dynamic Cascading Attacks:} In dynamic processes, the failure of one node triggers failures in others, constituting a complex dynamic propagation process. Consequently, the failure sequence must be obtained through real simulations or other efficient methods. Given that this study aims to rapidly and accurately assess robustness, the node failure sequence under this attack is assumed to be obtained as prior features.
    \end{enumerate}

\subsection{Hypergraph Isomorphism Network}
Inspired by GIN and HWL, we present the following lemma and theorem. The proofs of the first two can be found in the appendix while the proof of the third is provided in the paper on GIN \cite{xu2018powerful}:
\begin{lemma}
    \label{lemma1}
    Let $H_1$ and $H_2$ be any two non-isomorphic hypergraphs. If an HGNN $\mathcal{A}:\mathcal{H} \rightarrow \mathbb{R}^d$ maps $H_1$ and $H_2$ to different embeddings, the HWL test also decides $H_1$ and $H_2$ are not isomorphic.
\end{lemma}
This lemma elucidates the upper bound of the expressive power of hypergraph neural networks.

\begin{theorem}
    \label{the0rem1}
    Let $\mathcal{A}:\mathcal{H} \rightarrow \mathbb{R}^d$ be an HGNN. With a sufficient number of HGNN layers, $\mathcal{A}$ maps any hypergraphs $H_1$ and $H_2$ that the hypergraph Weisfeiler-Lehman test of isomorphism decides as non-isomorphic, to different embeddings if the following conditions hold:
    \begin{enumerate}
        \item $\mathcal{A}$ aggregates and updates node features iteratively with

\begin{equation}
    \label{eq:hgnn_update_rules}
    \left\{
    \begin{aligned} 
        h_e^{(k)} &= \phi_e \left( h_e^{(k-1)}, f_v \left( \{h_{v_i}^{(k-1)} : v_i \in \mathcal{N}_v(e)\} \right) \right) \\
        h_v^{(k)} &= \phi_v \left( h_v^{(k-1)}, f_e \left( \{h_{e_i}^{(k)} : e_i \in \mathcal{N}_e(v)\} \right) \right)
    \end{aligned}
    \right.
\end{equation}
        where the functions $f_e$ and $f_v$, which operate separately on multisets of hyperedges and nodes, and $\phi_e$ and $\phi_v$ are injective.

        \item $\mathcal{A}$'s hypergraph-level readout, which operates on the multiset of both node features $\{h_v^k\}$ and hyperedge features $\{h_e^k\}$, is injective.
    \end{enumerate}
\end{theorem}
\begin{lemma}

Assume $\mathcal{X}$ is countable. There exists a function f : $\mathcal{X} \rightarrow \mathbb{R}^n$ so that for infinitely many choices of $\epsilon$, including all irrational numbers, $h(c,X)=(1+\epsilon) \cdot f(c)+\sum_{x\in X}f(x)$ is unique for each pair $(c,X)$, where $c \in \mathcal{X}$ and $X\subset\mathcal{X}$ is a multiset of bounded size. Moreover, any function $g$ over such pairs can be decomposed as $g(c,X)=\varphi((1+\epsilon) \cdot f(c)+\sum_{x\in X}f(x))$
for some function $\varphi$ \cite{xu2018powerful}.

\end{lemma}
Theorem 1 establishes the conditions under which Hypergraph Neural Networks achieve the upper bound of topological expressiveness, while Lemma 3 provides insights for their concrete implementation.
\subsubsection{Node Initialization}

   To ensure that the subsequent summation aggregation remains injective over multisets, raw features typically require to be mapped via an MLP. It is worth noting that for the first iteration with one-hot inputs, the pre-mapping MLP is optional in this case, as the raw summation already constitutes an injective mapping:
    \begin{equation}
    h_v^{(1)} = \text{MLP}^{(0)} \left(h_v^{(0)}\right)
    \end{equation}
\subsubsection{Node-to-Hyperedge Aggregation}
    As shown in Eq. \eqref{eq:node_to_edge}, the representation of a hyperedge is derived from a synergistic fusion of its intrinsic features and the aggregated features of its constituent nodes. By incorporating a learnable scalar $\epsilon_e$, the model is capable of distinguishing the hyperedge's self-identity from the structural messages contributed by its nodes. Since the node features have been mapped, the summation operation maintains injectivity, thereby satisfying the functional requirements of $f_v$ as posited in Theorem 1. Furthermore, the interaction between the hyperedge feature $h_e$ and the node feature set $h_v$ is modeled through an MLP architecture, modeled as $f^{(k+1)} \circ \varphi^{(k)}$. Specifically, the inner function $\varphi^{(k)}$ ensures the injective aggregation of the set $\{h_e,h_v\}$, while the resulting embedding is mapped by the outer transformation $f^{(k+1)} $ to preserve injectivity during the subsequent summation phase. Consequently, the discriminative criteria for the injectivity of function $\phi_e$ in Theorem 1 are rigorously fulfilled.
\begin{equation}
    h_e^{(l+1)} = \text{MLP}_e^{(l)}\left((1 + \epsilon_e^{(l)}) \cdot h_e^{(l)} +\sum_{v_i \in \mathcal{N}_v(e)} h_{v_i}^{(l)}\right)
    \label{eq:node_to_edge}
\end{equation}
\subsubsection{Hyperedge-to-Node Update}
Symmetrically, as shown in Eq. \eqref{eq:edge_to_node}, the representation of a node is derived from a synergistic fusion of its intrinsic features and the aggregated features of its incident hyperedges. By incorporating a learnable scalar $\epsilon_v$, the model is empowered to distinguish the node's self-identity from the structural messages contributed by its hyperedges. Since the hyperedge features have been mapped, the summation operation maintains injectivity, thereby satisfying the functional requirements of $f_e$ as posited in Theorem 1. Furthermore, the interaction between the node feature $h_v$ and the hyperedge feature set $h_e$ is modeled through an MLP architecture, modeled as $f^{(k+1)} \circ \varphi^{(k)}$. Specifically, the inner function $\varphi^{(k)}$ ensures the injective aggregation of the set $\{h_v, h_e\}$, while the resulting embedding is mapped by the outer transformation $f^{(k+1)}$ to preserve injectivity during the subsequent summation phase. Consequently, the discriminative criteria for the injectivity of function $\phi_v$ in Theorem 1 are rigorously fulfilled.
\begin{equation}
    h_v^{(l+1)} = \text{MLP}_v^{(l)} \left( (1 + \epsilon_v^{(l)}) \cdot h_v^{(l)} + \sum_{e_i \in \mathcal{N}_e(v)} h_{e_i}^{(l+1)} \right)
    \label{eq:edge_to_node}
\end{equation}

\subsubsection{Hypergraph-Level Readout}
The node representations corresponding to subtree patterns and the hyperedge representations corresponding to hyperedge patterns exhibit superior generalization capabilities in the early iterations. As the number of iterations increases, these representations evolve to become more granular and encompass global structural information. To comprehensively capture this full spectrum of structural data, we utilize information from all model depths. Specifically, since the node and hyperedge features at each layer $l$ have been mapped via MLPs, the layer-wise summation operator remains injective over the multisets of features. Consequently, the global concatenation of these injective summaries across all layers $l \in \{1, \dots, L\}$ preserves the overall injectivity of the mapping. This architectural design ensures that the final hypergraph embedding $H_{\text{graph}}$ fulfills the second requirement of Theorem 1, rigorously aligning our network with the theoretical discriminative power of the hypergraph WL isomorphism test.
\begin{equation}
    H_{\mathcal{G}} = \text{CONCAT} \left[ \sum_{v \in V} h_v^{(k)}, \sum_{e \in E} h_e^{(k)} \,\bigg\vert\, k=1, 2, \dots, L \right]
    \label{eq:jk}
\end{equation}

\subsection{Cosine Annealing Learning Rate Schedule}
To escape local minima and achieve better generalization performance, we employ the Cosine Annealing learning rate scheduling strategy during the training phase. Unlike step-decay strategies that reduce the learning rate abruptly, cosine annealing adjusts the learning rate $\eta_t$ smoothly following a cosine function. This allows the optimizer to maintain a relatively high learning rate in the initial stages for rapid exploration and decay precisely in later stages for fine-grained convergence.

Specifically, the learning rate $\eta_t$ at the $t$-th epoch is computed as:
\begin{equation}
\eta_t = \eta_{min} + \frac{1}{2}(\eta_{max} - \eta_{min}) \left( 1 + \cos \left( \frac{T_{cur}}{T_{max}} \pi \right) \right),
\label{eq:cosine_annealing}
\end{equation}
where $\eta_{max}$ and $\eta_{min}$ denote the initial (maximum) and minimum learning rates, respectively. $T_{cur}$ indicates the current number of epochs since the last restart, and $T_{max}$ represents the total number of epochs for a half-cosine cycle. In our implementation, we utilize the AdamW optimizer, integrating this scheduling strategy to adaptively adjust the step size for parameter updates, thereby enhancing the training stability and final prediction accuracy.

\section{Experimental Investigation}
\label{sec:experiments}
\begin{table*}[h]

\centering
\caption{Performance Comparison of Different Algorithms.}
\setlength{\tabcolsep}{2pt}
\renewcommand{\arraystretch}{1.3}
\label{tab:results}
\begin{tabular}{|c|c|llllllll|}
\hline
\multicolumn{2}{|c|}{\multirow{2}{*}{\textbf{Dataset}}} &
  \multicolumn{8}{c|}{Algorithm} \\ \cline{3-10} 
\multicolumn{2}{|l|}{} &
  \multicolumn{1}{c|}{\textbf{HWL-HIN}} &
  \multicolumn{1}{c|}{HGNNs} &
  \multicolumn{1}{c|}{GIN-MAS} &
  \multicolumn{1}{c|}{NRL-GT} &
  \multicolumn{1}{c|}{ATTRP} &
  \multicolumn{1}{c|}{SPP-CNN} &
  \multicolumn{1}{c|}{KNN} &
  \multicolumn{1}{c|}{DT} \\
\hline
\multirow{6}{*}[-30pt]{\rotatebox[origin=c]{90}{\textbf{Static}}} & ER &
  \multicolumn{1}{c|}{\begin{tabular}[c]{@{}c@{}}\textbf{0.00282±0.00262} \\ (1)\end{tabular}} &
  \multicolumn{1}{c|}{\begin{tabular}[c]{@{}c@{}}\textbf{0.00303±0.00262} \\ (2)\end{tabular}} &
  \multicolumn{1}{c|}{\begin{tabular}[c]{@{}c@{}}0.00341±0.00354 \\ (3)\end{tabular}} &
  \multicolumn{1}{c|}{\begin{tabular}[c]{@{}c@{}}0.00402±0.00361 \\ (4)\end{tabular}} &
  \multicolumn{1}{c|}{\begin{tabular}[c]{@{}c@{}}0.03308±0.02783 \\ (6)\end{tabular}} &
  \multicolumn{1}{c|}{\begin{tabular}[c]{@{}c@{}}0.03302±0.02568 \\ (5)\end{tabular}} &
  \multicolumn{1}{c|}{\begin{tabular}[c]{@{}c@{}}0.04817±0.04177 \\ (8)\end{tabular}} &
  \multicolumn{1}{c|}{\begin{tabular}[c]{@{}c@{}}0.04652±0.05734 \\ (7)\end{tabular}} \\ \cline{2-10}
 & SF &
  \multicolumn{1}{c|}{\begin{tabular}[c]{@{}c@{}}\textbf{0.00114±0.00200} \\ (1)\end{tabular}} &
  \multicolumn{1}{c|}{\begin{tabular}[c]{@{}c@{}}\textbf{0.00117±0.00200} \\ (2)\end{tabular}} &
  \multicolumn{1}{c|}{\begin{tabular}[c]{@{}c@{}}0.00127±0.00220 \\ (3)\end{tabular}} &
  \multicolumn{1}{c|}{\begin{tabular}[c]{@{}c@{}}0.00145±0.00263 \\ (4)\end{tabular}} &
  \multicolumn{1}{c|}{\begin{tabular}[c]{@{}c@{}}0.00265±0.00304 \\ (6)\end{tabular}} &
  \multicolumn{1}{c|}{\begin{tabular}[c]{@{}c@{}}0.00216±0.00324 \\ (5)\end{tabular}} &
  \multicolumn{1}{c|}{\begin{tabular}[c]{@{}c@{}}0.00311±0.00405 \\ (8)\end{tabular}} &
  \multicolumn{1}{c|}{\begin{tabular}[c]{@{}c@{}}0.00295±0.00498 \\ (7)\end{tabular}} \\ \cline{2-10}
 & UF &
  \multicolumn{1}{c|}{\begin{tabular}[c]{@{}c@{}}\textbf{0.00287±0.00256} \\ (1)\end{tabular}} &
  \multicolumn{1}{c|}{\begin{tabular}[c]{@{}c@{}}0.00314±0.00282  \\ (2)\end{tabular}} &
  \multicolumn{1}{c|}{\begin{tabular}[c]{@{}c@{}}0.00433±0.00388 \\ (3)\end{tabular}} &
  \multicolumn{1}{c|}{\begin{tabular}[c]{@{}c@{}}0.00595±0.00602 \\ (4)\end{tabular}} &
  \multicolumn{1}{c|}{\begin{tabular}[c]{@{}c@{}}0.04133±0.03243 \\ (6)\end{tabular}} &
  \multicolumn{1}{c|}{\begin{tabular}[c]{@{}c@{}}0.04102±0.03326 \\ (5)\end{tabular}} &
  \multicolumn{1}{c|}{\begin{tabular}[c]{@{}c@{}}0.05497±0.07056 \\ (7)\end{tabular}} &
  \multicolumn{1}{c|}{\begin{tabular}[c]{@{}c@{}}0.07603±0.05086 \\ (8)\end{tabular}} \\ \cline{2-10}
 & WS &
  \multicolumn{1}{c|}{\begin{tabular}[c]{@{}c@{}}\textbf{0.00927±0.00830} \\ (1)\end{tabular}} &
  \multicolumn{1}{c|}{\begin{tabular}[c]{@{}c@{}}0.01001±0.00890 \\ (2)\end{tabular}} &
  \multicolumn{1}{c|}{\begin{tabular}[c]{@{}c@{}}0.01018±0.00853 \\ (3)\end{tabular}} &
  \multicolumn{1}{c|}{\begin{tabular}[c]{@{}c@{}}0.01021±0.01080 \\ (4)\end{tabular}} &
  \multicolumn{1}{c|}{\begin{tabular}[c]{@{}c@{}}0.05801±0.03482 \\ (6)\end{tabular}} &
  \multicolumn{1}{c|}{\begin{tabular}[c]{@{}c@{}}0.05746±0.03985 \\ (5)\end{tabular}} &
  \multicolumn{1}{c|}{\begin{tabular}[c]{@{}c@{}}0.06028±0.07044 \\ (7)\end{tabular}} &
  \multicolumn{1}{c|}{\begin{tabular}[c]{@{}c@{}}0.06754±0.08373 \\ (8)\end{tabular}} \\ \cline{2-10}
 & SBM &
  \multicolumn{1}{c|}{\begin{tabular}[c]{@{}c@{}}\textbf{0.00423±0.00401} \\ (1)\end{tabular}} &
  \multicolumn{1}{c|}{\begin{tabular}[c]{@{}c@{}}0.00502±0.00470 \\ (2)\end{tabular}} &
  \multicolumn{1}{c|}{\begin{tabular}[c]{@{}c@{}}0.00546±0.00420 \\ (3)\end{tabular}} &
  \multicolumn{1}{c|}{\begin{tabular}[c]{@{}c@{}}0.00605±0.00460 \\ (4)\end{tabular}} &
  \multicolumn{1}{c|}{\begin{tabular}[c]{@{}c@{}}0.05889±0.03081 \\ (6)\end{tabular}} &
  \multicolumn{1}{c|}{\begin{tabular}[c]{@{}c@{}}0.05869±0.03185 \\ (5)\end{tabular}} &
  \multicolumn{1}{c|}{\begin{tabular}[c]{@{}c@{}}0.07195±0.06682 \\ (7)\end{tabular}} &
  \multicolumn{1}{c|}{\begin{tabular}[c]{@{}c@{}}0.07775±0.09422 \\ (8)\end{tabular}} \\ \cline{2-10}
 & MIX &
  \multicolumn{1}{c|}{\begin{tabular}[c]{@{}c@{}}\textbf{0.00468±0.00672} \\ (1)\end{tabular}} &
  \multicolumn{1}{c|}{\begin{tabular}[c]{@{}c@{}}0.00522±0.00641  \\ (2)\end{tabular}} &
  \multicolumn{1}{c|}{\begin{tabular}[c]{@{}c@{}}0.00770±0.00860 \\ (3)\end{tabular}} &
  \multicolumn{1}{c|}{\begin{tabular}[c]{@{}c@{}}0.00783±0.00800 \\ (4)\end{tabular}} &
  \multicolumn{1}{c|}{\begin{tabular}[c]{@{}c@{}}0.06036±0.04313 \\ (6)\end{tabular}} &
  \multicolumn{1}{c|}{\begin{tabular}[c]{@{}c@{}}0.05980±0.04229 \\ (5)\end{tabular}} &
  \multicolumn{1}{c|}{\begin{tabular}[c]{@{}c@{}}0.10615±0.07861 \\ (8)\end{tabular}} &
  \multicolumn{1}{c|}{\begin{tabular}[c]{@{}c@{}}0.07315±0.09471 \\ (7)\end{tabular}} \\ \hline
\multirow{6}{*}[-30pt]{\rotatebox[origin=c]{90}{\textbf{Dynamic}}} & ER &
  \multicolumn{1}{c|}{\begin{tabular}[c]{@{}c@{}}\textbf{0.00046±0.00040} \\ (1)\end{tabular}} &
  \multicolumn{1}{c|}{\begin{tabular}[c]{@{}c@{}}\textbf{0.00046±0.00040} \\ (1)\end{tabular}} &
  \multicolumn{1}{c|}{\begin{tabular}[c]{@{}c@{}}0.00163±0.00140 \\ (2)\end{tabular}} &
  \multicolumn{1}{c|}{\begin{tabular}[c]{@{}c@{}}0.00172±0.00160 \\ (3)\end{tabular}} &
  \multicolumn{1}{c|}{\begin{tabular}[c]{@{}c@{}}0.01750±0.01168 \\ (5)\end{tabular}} &
  \multicolumn{1}{c|}{\begin{tabular}[c]{@{}c@{}}0.01731±0.01124 \\ (4)\end{tabular}} &
  \multicolumn{1}{c|}{\begin{tabular}[c]{@{}c@{}}0.02326±0.02100 \\ (7)\end{tabular}} &
  \multicolumn{1}{c|}{\begin{tabular}[c]{@{}c@{}}0.01998±0.02460 \\ (6)\end{tabular}} \\ \cline{2-10}
 & SF &
  \multicolumn{1}{c|}{\begin{tabular}[c]{@{}c@{}}\textbf{0.00028±0.00034} \\ (2)\end{tabular}} &
  \multicolumn{1}{c|}{\begin{tabular}[c]{@{}c@{}}\textbf{0.00026±0.00027} \\ (1)\end{tabular}} &
  \multicolumn{1}{c|}{\begin{tabular}[c]{@{}c@{}}0.00069±0.00062 \\ (3)\end{tabular}} &
  \multicolumn{1}{c|}{\begin{tabular}[c]{@{}c@{}}0.00094±0.00100 \\ (4)\end{tabular}} &
  \multicolumn{1}{c|}{\begin{tabular}[c]{@{}c@{}}0.00704±0.00519 \\ (7)\end{tabular}} &
  \multicolumn{1}{c|}{\begin{tabular}[c]{@{}c@{}}0.00674±0.00476 \\ (5)\end{tabular}} &
  \multicolumn{1}{c|}{\begin{tabular}[c]{@{}c@{}}0.00693±0.00845 \\ (6)\end{tabular}} &
  \multicolumn{1}{c|}{\begin{tabular}[c]{@{}c@{}}0.01085±0.01281 \\ (8)\end{tabular}} \\ \cline{2-10}
 & UF &
  \multicolumn{1}{c|}{\begin{tabular}[c]{@{}c@{}}\textbf{0.00049±0.00040} \\ (1)\end{tabular}} &
  \multicolumn{1}{c|}{\begin{tabular}[c]{@{}c@{}}\textbf{0.00053±0.00050} \\ (2)\end{tabular}} &
  \multicolumn{1}{c|}{\begin{tabular}[c]{@{}c@{}}0.00121±0.00090 \\ (3)\end{tabular}} &
  \multicolumn{1}{c|}{\begin{tabular}[c]{@{}c@{}}0.00162±0.00180 \\ (4)\end{tabular}} &
  \multicolumn{1}{c|}{\begin{tabular}[c]{@{}c@{}}0.01223±0.00796 \\ (6)\end{tabular}} &
  \multicolumn{1}{c|}{\begin{tabular}[c]{@{}c@{}}0.01205±0.00852 \\ (5)\end{tabular}} &
  \multicolumn{1}{c|}{\begin{tabular}[c]{@{}c@{}}0.02132±0.01466 \\ (8)\end{tabular}} &
  \multicolumn{1}{c|}{\begin{tabular}[c]{@{}c@{}}0.01410±0.01758 \\ (7)\end{tabular}} \\ \cline{2-10}
 & WS &
  \multicolumn{1}{c|}{\begin{tabular}[c]{@{}c@{}}\textbf{0.00043±0.00025} \\ (1)\end{tabular}} &
  \multicolumn{1}{c|}{\begin{tabular}[c]{@{}c@{}}\textbf{0.00046±0.00043} \\ (2)\end{tabular}} &
  \multicolumn{1}{c|}{\begin{tabular}[c]{@{}c@{}}0.00135±0.00112 \\ (3)\end{tabular}} &
  \multicolumn{1}{c|}{\begin{tabular}[c]{@{}c@{}}0.00157±0.00168 \\ (4)\end{tabular}} &
  \multicolumn{1}{c|}{\begin{tabular}[c]{@{}c@{}}0.00813±0.00723 \\ (5)\end{tabular}} &
  \multicolumn{1}{c|}{\begin{tabular}[c]{@{}c@{}}0.01545±0.01054 \\ (8)\end{tabular}} &
  \multicolumn{1}{c|}{\begin{tabular}[c]{@{}c@{}}0.00834±0.01101 \\ (6)\end{tabular}} &
  \multicolumn{1}{c|}{\begin{tabular}[c]{@{}c@{}}0.01171±0.01517 \\ (7)\end{tabular}} \\ \cline{2-10}
 & SBM &
  \multicolumn{1}{c|}{\begin{tabular}[c]{@{}c@{}}\textbf{0.00049±0.00050} \\ (2)\end{tabular}} &
  \multicolumn{1}{c|}{\begin{tabular}[c]{@{}c@{}}\textbf{0.00046±0.00040} \\ (1)\end{tabular}} &
  \multicolumn{1}{c|}{\begin{tabular}[c]{@{}c@{}}0.00116±0.00102 \\ (3)\end{tabular}} &
  \multicolumn{1}{c|}{\begin{tabular}[c]{@{}c@{}}0.00184±0.00169 \\ (4)\end{tabular}} &
  \multicolumn{1}{c|}{\begin{tabular}[c]{@{}c@{}}0.01494±0.00883 \\ (6)\end{tabular}} &
  \multicolumn{1}{c|}{\begin{tabular}[c]{@{}c@{}}0.01485±0.00879 \\ (5)\end{tabular}} &
  \multicolumn{1}{c|}{\begin{tabular}[c]{@{}c@{}}0.01752±0.01759 \\ (8)\end{tabular}} &
  \multicolumn{1}{c|}{\begin{tabular}[c]{@{}c@{}}0.01673±0.02081 \\ (7)\end{tabular}} \\ \cline{2-10}
 & MIX &
  \multicolumn{1}{c|}{\begin{tabular}[c]{@{}c@{}}\textbf{0.00063±0.00048} \\ (1)\end{tabular}} &
  \multicolumn{1}{c|}{\begin{tabular}[c]{@{}c@{}}\textbf{0.00075±0.00059} \\ (2)\end{tabular}} &
  \multicolumn{1}{c|}{\begin{tabular}[c]{@{}c@{}}0.00162±0.00132 \\ (3)\end{tabular}} &
  \multicolumn{1}{c|}{\begin{tabular}[c]{@{}c@{}}0.00248±0.00160 \\ (4)\end{tabular}} &
  \multicolumn{1}{c|}{\begin{tabular}[c]{@{}c@{}}0.03088±0.03001 \\ (6)\end{tabular}} &
  \multicolumn{1}{c|}{\begin{tabular}[c]{@{}c@{}}0.01424±0.01166 \\ (5)\end{tabular}} &
  \multicolumn{1}{c|}{\begin{tabular}[c]{@{}c@{}}0.03866±0.04020 \\ (8)\end{tabular}} &
  \multicolumn{1}{c|}{\begin{tabular}[c]{@{}c@{}}0.03258±0.04636 \\ (7)\end{tabular}} \\ \hline
\end{tabular}
\end{table*}
\subsection{Baseline Methods and Experimental Data}

In this section, we detail the experimental datasets, baseline methods, and implementation details.

To comprehensively evaluate the robustness prediction capability of HWL-HIN across diverse topological structures, we generated five distinct types of synthetic hypergraphs. Each dataset represents a unique structural property commonly found in real-world complex systems. All hypergraphs are generated with a fixed node size of $N=200$ to maintain consistency. The specific configurations are as follows:

\begin{itemize}
    \item Erd\H{o}s-R\'enyi (ER): Represents random hypergraphs where connections are formed stochastically. We set the connection probability $p=0.05$, ensuring the graph is sparse yet connected, serving as a baseline for unstructured connectivity.
    
    \item Watts-Strogatz (WS): Simulates small-world properties characterized by high clustering and short path lengths. We initialize a regular lattice where each node connects to $k_{nn}=10$ nearest neighbors, and then randomize edges with a rewiring probability $p_{rw}=0.5$.
    
    \item Scale-Free (SF): Generated via the preferential attachment mechanism to mimic power-law hyperdegree distributions. Each new node added to the network introduces $m=5$ new hyperedges, creating a topology with significant hubs.
    
    \item Stochastic Block Model (SBM): Models community structures typical of social networks. We configure $C=5$ distinct communities. To ensure strong modularity, the intra-community connection probability is set to $p_{in}=0.1$, which is an order of magnitude higher than the inter-community probability $p_{out}=0.01$.
    
    \item Uniform Random (UF): A variation of random hypergraphs designed to test sensitivity to hyperedge size. Unlike other datasets where hyperedge sizes vary, we strictly fix the cardinality of every hyperedge to $k=5$.
\end{itemize}

Dataset Construction and Splitting: 
We constructed six distinct datasets for training and evaluation. The first five datasets ($\mathcal{D}_{ER}, \mathcal{D}_{WS}, \mathcal{D}_{SF}, \mathcal{D}_{SBM}, \mathcal{D}_{UF}$) are homogeneous, containing only one type of topology. For these, we generated 1,200 samples each, split into 1,000 for training and 200 for testing. 
To further assess the generalization ability of our surrogate model across unseen or hybrid topologies, we constructed a sixth Mixed Dataset ($\mathcal{D}_{Mix}$). This dataset aggregates 500 samples from each of the five generative models, resulting in a comprehensive training set of 2,500 samples. An independent set of 200 mixed samples was reserved for testing.

\subsection{Baseline Methods}
\label{subsec:baselines}

To strictly evaluate the proposed HWL-HIN, we compare it with graph-based methods operating on the transformed bipartite incidence graphs. The baselines include a strict control group model HGNNs, representative GIN-MAS \cite{wu2024multitask} and NRL-GT \cite{zhang2024graph}, matrix-based deep learning methods SPP-CNN \cite{wu2023spp} and ATTRP \cite{huang2021unignn}, as well as classic machine learning algorithms KNN and DT. Specifically, for graph-based models GIN-MAS and NRL-GT, we slightly adjusted their optimization strategy to better accommodate the bipartite structure for enhanced training stability. For matrix-based methods SPP-CNN and ATTRP, the input resolution was set to $1028 \times 1028$. Aside from these specific adjustments, all other hyperparameters were set in accordance with the original papers to ensure a fair comparison.

\subsection{Performance Comparison and Analysis}
\label{subsec:results_analysis}

Table \ref{tab:results} presents a quantitative comparison of the proposed method against baselines. In the table, values denote the \textit{mean error} $\pm$ \textit{standard deviation}, and the subscripts indicate the average rank of each algorithm across datasets. The best-performing methods are highlighted in bold, and statistical significance is verified via a paired t-test with a significance level of $0.1$.

Overall, HWL-HIN demonstrates superior robustness prediction capabilities. Specifically, in static attack scenarios, HWL-HIN significantly outperforms the strict control group HGNNs. This advantage stems from HWL-HIN's strictly injective architecture, which provides higher expressive power to capture subtle structural distinctions compared to the standard aggregation used in HGNNs. However, under dynamic cascading attacks, the performance similarity between HWL-HIN and HGNNs increases. This is because the input features (i.e., the dynamic failure status) serve as strong predictors in this context; when the model relies heavily on these strong input features, its dependence on topological features decreases, thereby narrowing the gap between the two architectures.

Comparing with graph-based deep learning methods, GIN and NRL-GT perform competitively in static scenarios but suffer noticeable degradation in dynamic settings. This decline highlights a fundamental limitation in the baseline strategy: converting hypergraphs to bipartite incidence graphs inevitably incurs information loss, particularly regarding higher-order correlations essential for modeling complex cascading dynamics. While graph models can approximate static connectivity, they lack the native context to effectively track dynamic failure propagation on hyperedges. Furthermore, matrix-based methods (SPP-CNN, ATTRP) and traditional algorithms (KNN, DT) exhibit suboptimal performance across tasks. This is primarily due to the "curse of dimensionality" introduced by the bipartite conversion, which significantly increases the scale and sparsity of the input representation. Such high-dimensional, sparse structures pose a challenge for convolutional kernels and traditional classifiers, preventing them from effectively learning the underlying patterns of topological robustness.
\begin{table*}[htbp]
  \centering
  \caption{Ablation Study Results}
  \label{tab:ablation_study}
  \setlength{\tabcolsep}{4pt}
  \begin{tabular}{llcccccc}
    \toprule
    \textbf{Attack} & \textbf{Model} & \textbf{Original} & \textbf{Hyperdegree} & \textbf{Cardinality} & \textbf{Failure Order} & \textbf{Full Ablation} & \textbf{Learning Rate} \\
    \midrule
    \multirow{2}{*}{static}
    & HWL-HIN & \textbf{0.00470 ± 0.00680} & 0.00499 ± 0.00700 & 0.00488 ± 0.00665 & 0.00523 ± 0.00720 & 0.00529 ± 0.00700 & 0.00686 ± 0.00775 \\
    & HGNNs & \textbf{0.00522 ± 0.00641} & 0.00610 ± 0.00677 & 0.00615 ± 0.00729 & 0.00646 ± 0.00863 & * & 0.00921 ± 0.01017 \\
    \midrule
    \multirow{2}{*}{dynamic}
    & HWL-HIN & \textbf{0.00065 ± 0.00052} & 0.00076 ± 0.00062 & 0.00088 ± 0.00070 & 0.00470 ± 0.00383 & 0.00474 ± 0.00378 & 0.00222 ± 0.00155 \\
    & HGNNs & \textbf{0.00075 ± 0.00059} & 0.00082 ± 0.00062 & 0.00076 ± 0.00060 & 0.00473 ± 0.00379 & * & 0.00195 ± 0.00178 \\
    \bottomrule
  \end{tabular}
\end{table*}
\begin{figure}[!t]
    \centering
    \includegraphics[width=\linewidth]{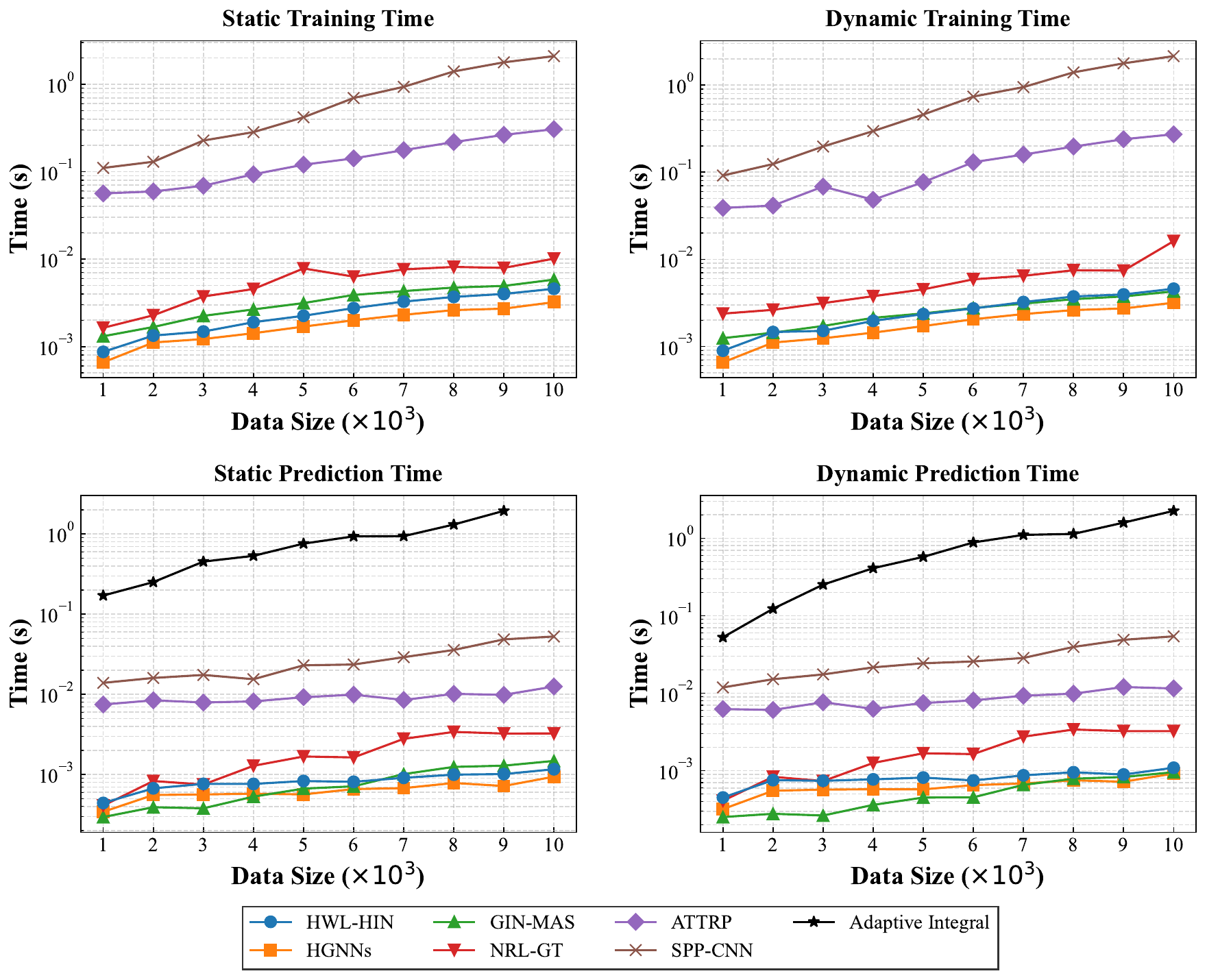}
    
    \caption{Efficiency comparison of different algorithms.}
    \label{fig:combined_results}
\end{figure}

\subsection{Computational Efficiency Analysis}
\label{subsec:time_efficiency}

We further evaluate the computational efficiency of different models as the scale of the hypergraph increases. The results indicate that the inference latency of hypergraph-based methods (HWL-HIN, HGNNs) and graph-based baselines (GIN-MAS, NRL-GT) remains comparable, showing negligible differences in time consumption. Notably, the strict control group, HGNNs, achieves slightly faster inference speeds than HWL-HIN, primarily attributable to its reduced parameter count resulting from the simplified non-injective aggregation mechanism. In stark contrast, matrix-based methods (SPP-CNN, ATTRP) exhibit significantly higher computational costs even without increasing the input image resolution. Furthermore, when the input resolution is scaled up to accommodate larger topological structures, these methods face severe scalability bottlenecks, frequently leading to GPU memory exhaustion (OOM), which renders them impractical for large-scale hypergraph tasks. Ultimately, regarding the overall efficacy of data-driven approaches, all examined machine learning surrogate models demonstrate a dramatic speed advantage over the adaptive integration method used for ground truth generation. Specifically, the prediction process is accelerated by approximately hundreds of times, highlighting the immense potential and efficiency of machine learning approaches in solving complex hypergraph robustness problems.

\subsection{Ablation Study}
\label{sec:ablation}

To ensure a rigorous comparison between HWL-HIN and the baseline HGNNs, we first address the structural asymmetry in their inputs and readout mechanisms. Specifically, HWL-HIN explicitly utilizes the node count within hyperedges as an additional input feature and employs a dual-channel readout that includes hyperedge embeddings. To align the experimental settings, we conducted an ablation by removing these specific components from HWL-HIN. In the context of the prediction robustness task, hyperedge features do not possess inherent independence as they are effectively derived from the aggregation of their constituent node features. Consequently, the explicit modeling of hyperedge-specific features yields limited marginal utility. Our empirical results support this, as ablating these components results in minimal performance fluctuations for HWL-HIN, which can be regarded as negligible. This confirms that the simplified model maintains representational integrity, thereby ensuring the fairness of the subsequent comparisons.

In static attack scenarios, the ablation results underscore a critical distinction in feature dependence. While removing input features causes a sharp performance drop for HGNNs, HWL-HIN maintains robust performance with minimal degradation. Most notably, when all input features are strictly ablated (i.e., learning solely from adjacency information), HWL-HIN retains its learning capability, whereas HGNNs fail to converge entirely. This stark contrast proves that HGNNs rely heavily on explicit node features to make predictions, while HWL-HIN possesses the unique ability to comprehend and learn directly from the underlying topological structure.

In dynamic cascading scenarios, the performance gap narrows. This is attributed to the "Failure Sequence" serving as an overwhelmingly strong prior feature; its dominance minimizes the marginal contribution of other features, leading to similar performance across models. However, consistent with static results, once all features (including the failure sequence) are ablated, HGNNs again lose the learning capability, while HWL-HIN remains functional. Finally, the ablation of the cosine annealing scheduler demonstrates that dynamically adjusting the learning rate throughout the iteration process significantly facilitates convergence and enhances stability for both attack patterns.
\section{Conclusion}
This paper proposes a Hypergraph Isomorphism Network method, which is proven to possess strictly equivalent expressive power to hypergraph WL isomorphism tests. This approach accelerates robustness analysis for higher-order networks modeled as hypergraphs. Experimental results demonstrate that deep learning prediction methods achieve hundreds of times faster computation than adaptive integration methods while maintaining high accuracy. We observe that HWL-HIN exhibits significant advantages when prediction tasks rely on topological structure, while their performance aligns with conventional hypergraph neural networks when input features dominate. Graph neural networks demonstrate effectiveness for hypergraph tasks, yet struggle to capture higher-order interactions between nodes and hyperedges, resulting in suboptimal performance. Moving forward, we will deploy HWL-HIN to tackle more intricate hypergraph-level tasks, confident that they will demonstrate more pronounced advantages over conventional hypergraph neural networks.

\appendices
\section{Proof of Lemma \ref{lemma1}}
Suppose after $i$ iterations, an HGNN $\mathcal{A}$ has $\mathcal{A}(H_1)\neq\mathcal{A}(H_2)$ but the HWL test cannot decide $H_1$ and $H_2$ are non-isomorphic. Therefore $H_1$ and $H_2$ always have the same collection of node labels from iteration $0$ to $k$ in the HWL test.
We next demonstrate that for two hypergraphs $H_1$ and $H_2$, if the HWL node labels are $l_v^{(i)}=l_u^{(i)}$ and the hyperedge labels are $l_e^{(i)}=l_\omega^{(i)}$, then the HGNN node features and hyperedge features are consistently $h_v^{(i)}=h_u^{(i)}$ and $h_e^{(i)}=h_\omega^{(i)}$ for any iteration $i$. For the base case $i=0$, this holds trivially due to the identical input features. Assuming the hypothesis holds for the $i$-th iteration, if for any $u$,  $v$, $e$ and $\omega$, $l_v^{(i+1)}=l_u^{(i+1)}$ and $l_e^{(i+1)}=l_\omega^{(i+1)}$, then it must be the case that:
    \begin{equation}
                (l_e^{(i)},\{l_v^{(i)}:v\in \mathcal{N}_v(e)\})=(l_\omega^{(i)},\{l_u^{(i)}:u \in \mathcal{N}_v(\omega) \})
    \end{equation}
    By our assumption, we must have:
    \begin{equation}
                (h_e^{(i)},\{h_v^{(i)}:v\in \mathcal{N}_v(e)\})=(h_\omega^{(i)},\{h_u^{(i)}:u \in \mathcal{N}_v(\omega) \})
    \end{equation}
Analogous to GNNs, HGNNs employ identical AGGREGATE and COMBINE functions. Since identical inputs yield identical outputs, we obtain $h_e^{(i+1)}=h_\omega^{(i+1)}$. Consequently, we have:
    \begin{equation}
                (h_v^{(i)},\{h_e^{(i+1)}:e\in\mathcal{N}_e(v)\})=(h_u^{(i)},\{h_\omega^{(i+1)}:\omega \in \mathcal{N}_e(u)\})
    \end{equation}
Similarly, we obtain $h_v^{(i+1)}=h_u^{(i+1)}$. Therefore, by induction, there exist two valid mappings $\phi_e$ and $\phi_v$ for any $v, e \in H$ such that:
    \begin{equation}
               \left\{
               \begin{aligned}
                &h_v^{(i)}=\phi_v(l_v^{(i)})\\
                &h_e^{(i)}=\phi_e(l_e^{(i)})
               \end{aligned}
               \right.
    \end{equation}
    It follows from the fact that $H_1$ and $H_2$ have the same multiset of HWL neighborhood labels that $H_1$ and $H_2$ also have the same collection of both the node and hyperedge neighborhood features.
    \begin{equation}
                \{h_e^{(i)},\{h_v^{(i)}:v\in \mathcal{N}_v(e)\}\}=\{\phi_e(l_e^{(i)}),\{\phi_v(l_v^{(i)}):v\in \mathcal{N}_v(e)\}\}
    \end{equation}
    Thus, $h_e^{(i+1)}$ are the same, then we have:
    \begin{equation}
                \small
                \{h_v^{(i)},\{h_e^{(i+1)}:e\in\mathcal{N}_e(v)\}\}=\{\phi_v(l_v^{(i)}),\{\phi_e(l_e^{(i+1)}):e\in\mathcal{N}_e(v)\}\}
    \end{equation}
    Similarly, $h_v^{(i+1)}$ are the same.
    Consequently, we have the same collection of node features and hyperedge features for $H_1$ and $H_2$. Given that the hypergraph readout function is permutation invariant over the multiset of node and hyperedge features, the resulting representation is independent of the input ordering, $\mathcal{A}(H_1)=\mathcal{A}(H_2)$. Hence we have reached a contradiction.
\hfill $\square$

\section{Proof of Theorem \ref{the0rem1}}
Let $\mathcal{A}$ be an HGNN where the condition holds. Let $H_1$, $H_2$ be any hypergraphs which the HWL test decides as non-isomorphic at iteration $K$. Because the hypergraph-level readout function is injective, i.e., it maps distinct multisets of both node features and hyperedge features into unique embeddings, it suffices to show that $\mathcal{A}$’s neighborhood aggregation process, with sufficient iterations, embeds $H_1$ and $H_2$ into different multisets of both node features and edge features. Let us assume $\mathcal{A}$ updates node and hyperedge representations as:
\begin{equation}
    \label{eq:hgnn_update_rules}
    \left\{
    \begin{aligned} 
        h_e^{(k)} &= \phi_e \left( h_e^{(k-1)}, f_v \left( \{h_{v_i}^{(k-1)} : v_i \in \mathcal{N}_v(e)\} \right) \right) \\
        h_v^{(k)} &= \phi_v \left( h_v^{(k-1)}, f_e \left( \{h_{e_i}^{(k)} : e_i \in \mathcal{N}_e(v)\} \right) \right)
    \end{aligned}
    \right.
\end{equation}
where the functions $f_e$, $f_v$, $\phi_e$ and $\phi_v$ are injective. The HWL test applies a predetermined injective hash function $g$ to update the HWL edge labels and the node labels:
\begin{equation}
    \left\{
    \begin{aligned} 
        l_e^{(k)} &= g \left( l_e^{(k-1)},  \{l_{v_i}^{(k-1)} : v_i \in \mathcal{N}_v(e) \} \right)  \\
        l_v^{(k)} &= g \left( l_v^{(k-1)},  \{l_{e_i}^{(k)} : e_i \in \mathcal{N}_e(v)\} \right) 
    \end{aligned}
    \right.
\end{equation}
We use induction to prove that for any iteration $k$, there always exist two injective functions $\varphi_e$ and $\varphi_v$ such that:
     \begin{equation}
            \left\{
            \begin{aligned} 
                h_e^{(k)}=\varphi_e(l_e^{(k)})  \\
                h_v^{(k)}=\varphi_v(l_v^{(k)})
            \end{aligned}
            \right.
    \end{equation}
This apparently holds for $k=0$ because the initial node features and hyperedge features are the same for HWL and $\mathcal{A}$ $h_e^{(0)}=l_e^{(0)}$ and $h_v^{(0)}=l_v^{(0)}$ for all $v,e\in H_1,H_2$. Thus, $\varphi_e$ and $\varphi_v$ could be the identity function for $k = 0$. Suppose this holds for iteration $k-1$; we show that it also holds for $k$. By substituting $h_e^{(k-1)}$ with $\varphi_e(l_e^{(k-1)})$ and $h_v^{(k-1)}$ with $\varphi_v(l_v^{(k-1)})$ and deducing:
\begin{equation}
    \begin{aligned}
        h_e^{(k)} &= \phi_e \left( h_e^{(k-1)}, f_v \left( \{h_{v_i}^{(k-1)} : v_i \in \mathcal{N}_v(e)\} \right) \right)\\
        &= \phi_e \left( \varphi_e(l_e^{(k-1)}), f_v \left( \left\{ \varphi_v(l_{v_i}^{(k-1)}) : v_i \in \mathcal{N}_v(e) \right\} \right) \right) \\
          &= \psi_e \left(l_e^{(k-1)},\{l_{v_i}^{(k-1)} : v_i \in \mathcal{N}_v(e)\}  \right) \\
          &= \psi_e \circ g^{-1}(l_e^{(k)})\\
          &= \varphi_e (l_e^{(k)})
    \end{aligned}
\end{equation}
where $\psi_e$ and $\varphi_e=\psi_e \circ g^{-1}$ are injective, since the composition of injective functions is injective. Then, substituting  $h_v^{(k-1)}$ with $\varphi_v(l_v^{(k-1)})$  and  $h_e^{(k)}$ with $\varphi_e(l_e^{(k)})$, We deduce:
\begin{equation}
    \begin{aligned}
        h_v^{(k)} &= \phi_v \left( h_v^{(k-1)}, f_e \left( \{h_{e_i}^{(k)} : e_i \in \mathcal{N}_e(v)\} \right)\right)\\
        &=  \phi_v \left( \varphi_v(l_v^{(k-1)}), f_e \left( \{\varphi_e(l_{e_i}^{(k)}) : e_i \in \mathcal{N}_e(v)\} \right)\right) \\
          &= \psi_v \left(l_v^{(k-1)},\{l_{e_i}^{(k)} : e_i \in \mathcal{N}_e(v)\}  \right) \\
          &= \psi_v \circ g^{-1}(l_v^{(k)})\\
          &= \varphi_v (l_v^{(k)})
    \end{aligned}
\end{equation}
where $\psi_v$ and $\varphi_v=\psi_v\circ g^{-1}$ are injective. At the $K$-th iteration, the HWL test decides that $H_1$ and $H_2$ are non-isomorphic, that is the multisets $l_v^{(k)}$ and $l_e^{(k)}$ are different from $H_1$ and $H_2$. The hypergraph neural network $\mathcal{A}$’s node embeddings and hyperedge embeddings must be different from $H_1$ and $H_2$ because of the injectivity of $\varphi_e$ and $\varphi_v$.
\hfill $\square$
\bibliographystyle{IEEEtran} 
\bibliography{main}

\vfill

\end{document}